\documentclass[letterpaper, 10 pt, conference]{ieeeconf}
\IEEEoverridecommandlockouts                              % This command is only
\renewcommand{\baselinestretch}{0.97}
\usepackage[usenames,dvipsnames]{xcolor}
\usepackage{amsmath,amsthm}
\usepackage{bm}
\usepackage{amssymb}
\usepackage{mathtools}
\usepackage{mathabx}
\usepackage{dsfont}
\usepackage{graphicx}
\usepackage{epstopdf}
\usepackage[noadjust]{cite}
\usepackage{etoolbox}
\usepackage{filecontents}
\usepackage{tikz}
\usepackage{scalerel}
\usepackage[ruled,vlined]{algorithm2e}
\usepackage{subcaption}
\usepackage{float}
\restylefloat{table}
\usepackage{tabularx}

\newtheorem{theorem}{Theorem}

\newtheorem{lemma}{Lemma}

\allowdisplaybreaks

\title{\LARGE \bf
Learning Latent Fractional dynamics with Unknown Unknowns}

\author{Gaurav~Gupta$^{\dag}$ $\quad$ S\'ergio~Pequito$^{\ddagger}$  $\quad$  Paul~Bogdan$^{\dag}$
\thanks{
$^{\dag}$Ming Hsieh Department of Electrical and Computer Engineering, University of Southern California, Los Angeles, CA, USA {\tt\small \{ggaurav,pbogdan\}@usc.edu}}
\thanks{
$^{\ddagger}$Department of Industrial and Systems Engineering, Rensselaer Polytechnic Institute, Troy, NY, USA
{\tt\small goncas@rpi.edu}}
}

\begin{document}

\maketitle
\thispagestyle{empty}
\pagestyle{empty}

%
%----------------------------------------------------------------------------
%-----------------------------------ABSTRACT---------------------------------
%----------------------------------------------------------------------------
%
\begin{abstract}
Despite significant effort in understanding complex systems (CS), we lack a theory for modeling, inference, analysis and efficient control of time-varying complex networks (TVCNs) in uncertain environments. From brain activity dynamics to microbiome, and even chromatin interactions within the genome architecture, many such TVCNs exhibits a pronounced spatio-temporal fractality. Moreover, for many TVCNs only limited information (e.g., few variables) is accessible for modeling, which hampers the capabilities of analytical tools to uncover the true degrees of freedom and infer the CS model, the hidden states and their parameters. Another fundamental limitation is that of understanding and unveiling of unknown drivers of the dynamics that could sporadically excite the network in ways that straightforward modeling does not work due to our inability to model non-stationary processes. Towards addressing these challenges, in this paper, we consider the problem of learning the fractional dynamical complex networks under unknown unknowns (i.e., hidden drivers) and partial observability (i.e., only partial data is available). More precisely, we consider a generalized modeling approach of TVCNs consisting of discrete-time fractional dynamical equations and propose an iterative framework to determine the network parameterization and predict the state of the system. We showcase the performance of the proposed framework in the context of task classification using real electroencephalogram data.

\end{abstract}
% 

%
%----------------------------------------------------------------------------
%-----------------------------SEC-I : INTRODUCTION---------------------------
%----------------------------------------------------------------------------
%
\section{Introduction}
\label{sec:intro}

Time-varying complex networks (TVCNs) provides a comprehensive mathematical framework for modeling complex biological, social and technological systems such as the brain dynamic activity \cite{xue1,gauravACC2018,gauravICCPS2018}, the gene expression and interactions \cite{wang2016NatureComm, lopes2011Gene}, the bacteria dynamics \cite{Arafa2013,rihan2014fractional,Koorehdavoudi}, or the swarm robotics \cite{Couceiro2012,COUCEIRO201636}. Many such time varying complex (biological) networks exhibit complex spatio-temporal interactions. For instance, the short- and long-range interactions among neurons contribute to the emergence of long-range memory and fractional dynamics at macroscopic brain regions. Moreover, the non-stationarity which arises in most of the bio-physical processes require modeling techniques supporting interaction among variables in space and time. A computationally efficient strategy for constructing compact yet accurate mathematical models of TVCNs relies on describing the self-activity of nodes in TVCNs through fractional order operators \cite{Moon,Lundstrom,Werner,Turcott,Thurner}.

%Time-varying complex networks (TVCNs) exhibiting interactions across nodes through spatial correlations, and time-dependence from self-activities are considered. The models for TVCNs with fractional dynamics capture long-range time dependence of node self-activity by the fractional operator \cite{Moon,Lundstrom,Werner,Turcott,Thurner}. The accurate and compact modeling of real-world systems has always been a challenge, but despite of that the applications span a large domain ranging from brain dynamic activity \cite{xue1,gauravACC2018,gauravICCPS2018}, gene interactions \cite{wang2016NatureComm, lopes2011Gene}, bacteria dynamics \cite{Arafa2013,rihan2014fractional,Koorehdavoudi}, swarm robotics \cite{Couceiro2012,COUCEIRO201636}. The non-stationarity which arises in most of the physical process require modeling techniques supporting interaction among variables in space and time.

The modeling of interactions across nodes require assumption of complete knowledge of the associated complex network (CN). However, due to the experimental limitations, (e.g., resource constraint, limiting probing capabilities\footnote{As Wigner argued once: `It is the skill and ingenuity of the experimenter which show him phenomena which depend on a relatively narrow set of relatively easily realizable and reproducible conditions' \cite{wigner}.}), only a part of the complete CN is available at most of the times. The influence of so-called latent nodes on the observed nodes can be captured by noise, which may not be a good approach as these latent nodes activities have specific patterns. For example, a sensor capturing brain region activity (known to be fractional dynamic), going to be relaxed in the future, and thereby making it latent. In the current work, we are concerned with fractional dynamical model partly because of their celebrated success in modeling several physiological signals like electroencephalogram (EEG), electrocardiogram (ECG), electromyogram (EMG), and blood oxygenation level dependent (BOLD)~\cite{magin2012fractional,baleanu2011fractional}. In experimental setups, for example the brain, the recorded data is influenced by subcortical regions which are not probed. Instead of labeling them as unknown inputs to the observed system, it can be realized that they also obey the fractional dynamical patterns. In other situations, it is often required to relax some sensors, due to power limitations or sensor failure. In all such conditions, we have only partial information regarding the complete TVCN, and we wish to better predict the observed data in the presence of latent fractional dynamics as well as unknown drivers.

There is rich literature regarding study of networks with latent nodes. To list a few, the importance of realizing the inclusion of latent nodes in the context of linear time invariant (LTI) systems has been explored in \cite{JalaliICML2012, AnimaICML2013,Geiger2015}, for Bayesian networks in  \cite{elidanJMLR2007}, and graphical models with Gaussian distribution of nodes in \cite{chandrasekaran2012}, complex systems \cite{matteo2013ComplexSys}. The Markovian assumption helps in the case of LTI systems, and in some sense to uncover the latent nodes. However, the LTI systems are not sufficient to accurately model physiological signals, such as EEG, ECG and BOLD (just to mention a few) due to their inability in capturing the long-range memory property of the biological signals. To the best of authors' knowledge, the latent node framework is still non-existent in the context of discrete-time fractional dynamical systems. The closest work concerning fractional-order systems are \cite{xue1} and \cite{gauravACC2018}, where complete knowledge of the CN is assumed. The work in \cite{gauravACC2018} has generalized the work in \cite{xue1}, and included the contribution of unknown drivers in the system, but also with the assumption of complete knowledge of the nodes in the TVCN. The unknown unknowns were designated as external activities which does not obey the fractional dynamics of the CN.

Our contributions in this work are as follows: (i) we present a TVCN having fractional dynamics with latent nodes; (ii) jointly estimate the fractional latent node activities, and unknown drivers which does not obey fractional dynamics from observed data; (iii) iteratively estimate the complete model (latent $+$ observed) parameters in the likelihood sense.

The rest of the paper is organized as follows. Section \ref{sec:probForm} introduces the system model with latent nodes considered in this paper, and then formally presents the main problem statement. In Section\,\ref{sec:ModelEst}, we provide solutions to the considered problems as the main results, and in Section\,\ref{sec:experi}, we evaluate the proposed methods on simulated and real-world datasets. Finally, we conclude the work and present future directions in Section\,\ref{sec:concl}.
%
%
%----------------------------------------------------------------------------
%-------------------------SEC-II : PROBLEM FORMULATION-----------------------
%----------------------------------------------------------------------------
%

\section{Problem Formulation}
\label{sec:probForm}

We begin by introducing in Section\,\ref{ssec:sysModel} the TVCN system model obeying fractional-order dynamical evolution with latent nodes and unknown excitations. Next, in Section\,\ref{ssec:sysId}, we describe the main problem addressed in this paper.

\subsection{System Model}
\label{ssec:sysModel}

We consider a time-varying complex network (TVCN) described by a linear discrete-time fractional-order model with latent nodes, which can be mathematically written as

\begin{eqnarray}
\Delta^{\alpha}\begin{bmatrix}
x[k+1]\\
z[k+1]
\end{bmatrix} &=& \begin{bmatrix}
A_{11} & A_{12} \\
A_{21} & A_{22}
\end{bmatrix}
\begin{bmatrix}
x[k]\\
z[k]
\end{bmatrix}\nonumber\\
&&\quad +\begin{bmatrix}
B_{1}\\
B_{2}
\end{bmatrix}u[k] + \begin{bmatrix}
e_{1}[k]\\
e_{2}[k]
\end{bmatrix},
\label{eqn:sysModel}
\end{eqnarray}

\noindent where $x\in \mathbb{R}^{n}$ are the observed state variables, $z \in \mathbb{R}^{m}$ are the latent states, and $u \in \mathbb{R}^{p}$ are the unknown excitations. The system matrices $(\alpha, A_{11}, A_{12}, A_{21}, A_{22}, B_{1}, B_{2})$ are of appropriate dimensions. The noise variables are assumed to be uncorrelated across observed and latent nodes with $e_{1}\sim \mathcal{N}(0, \Sigma_{1})$ and $e_{2}\sim \mathcal{N}(0, \Sigma_{2})$. The fractional-order derivative in equation (\ref{eqn:sysModel}) obeys the discrete form for every node, either observed or latent, as follows \cite{oldham2006fractional}:
\begin{equation}
\begin{split}
\Delta^{\alpha}x[k] &= \sum\limits_{j=0}^{k}\Psi_{j}^{1}x[k-j],\\
\Delta^{\alpha}z[k] &= \sum\limits_{j=0}^{k}\Psi_{j}^{2}z[k-j],
\end{split}
\label{eqn:fracExpan}
\end{equation}
\noindent where the matrices $\Psi_{j}^{1} = \text{diag} ( \psi(\alpha_{1}^{o},j),\hdots,\psi(\alpha_{n}^{o},j))$ and  $\Psi_{j}^{2} = \text{diag}(\psi(\alpha_{1}^{l},j),\hdots,\psi(\alpha_{m}^{l},j))$ with $\psi(\alpha,j) = \frac{\Gamma(j-\alpha)}{\Gamma(-\alpha)\Gamma(j+1)}$, and $\Gamma(.)$ denotes the gamma function. The fractional-order coefficients corresponding to the $i$th node of observed and latent variables are denoted by $\alpha_{i}^{o}$ and $\alpha_{i}^{l}$, respectively.

%Next, we formally state the problem of model estimation when we observe $x$ and some part of the network, i.e. $z$, is hidden. Also, we jointly solve the problem of estimating unknown excitations which does not follow the fractional dynamical behavior.

\subsection{System Identification with Latent nodes}
\label{ssec:sysId}

The physiological signals, e.g. EEG and ECG, display spatio-temporal behavior, where the temporal component shows long-range memory dependence as realized in \cite{xue1, gauravACC2018}. As a consequence, properly modeled by the systems described in Section\,\ref{ssec:sysModel}. When trying to obtain the system representation, i.e., to identify the system's parameters, the model estimation is contingent on the complete knowledge of the assumed complex-network dynamics; specifically, their nodes' activities in terms of time-series. Such assumptions may not hold in most of the cases where we are provided only with partial data. In such cases, to better predict the complex systems dynamics, it is beneficial to incorporate latent nodes. In this regard, the problem considered in this work can be stated as follows.

\textbf{Problem Statement:} \textit{Given} the partial data (observed) $x[k]$ in terms of \mbox{time-series} across a time-horizon $k\in \{1,\hdots,N\}$, and knowledge of the fractional orders of latent nodes $\alpha_{i}^{l} (1\leq i\leq m)$. \textit{Estimate} the model parameters $(\alpha^{o}, \allowbreak A_{11}, \allowbreak A_{12}, \allowbreak A_{21}, \allowbreak A_{22}, \allowbreak B_{1}, \allowbreak B_{2})$ and latent states $\{z[k]\}_{1}^{N-1}$, and the unknown inputs $\{u[k]\}_{1}^{N-1}$.

This problem will build upon the models in \cite{gauravACC2018,xue1}, but with striking difference of availability of only the partial data, and presence of latent nodes in the model. In contrast to \cite{gauravACC2018}, this work also relax the assumption of knowledge of the input matrices, and they will be computed as part of the system's parameters. Notice that we are implicitly assuming that the fractional-order coefficients are constant over time since these have been shown to be empirically slowly time-varying. We will see in the Section\,\ref{sec:experi} that by considering latent nodes we can improve the prediction accuracy of the observed data. In the next section, we detail the assumptions required, and solution to this estimation problem.

\section{Latent Model Estimation}
\label{sec:ModelEst}

Due to notational convenience, we denote $x[k]$ as $x_{k}$. The model estimation procedure begins with, first making an estimate of the latent node activities with assumptions that some approximation of the system's parameters are known. A fractional Kalman filtering approach similar to \cite{Sierociuk06fractionalkalman} under Bayesian approximation is used for this purpose in Section\,\ref{ssec:kalmanFilt}. Subsequently, we will use the new data (estimated latent and the available data) to perform the system identification. In Section\,\ref{ssec:mle}, we present an iterative algorithm to jointly estimate the system's parameters and the unknown unknowns.

%The model estimation procedure begins with the definition of conditional distribution of the latent nodes activities $z_{k}$ with respect to the observations $x_{k}$. A fractional-order Kalman filtering approach similar to \cite{Sierociuk06fractionalkalman} will be used. Building upon these distributions, we seek to optimize maximum likelihood estimation (MLE) of the system model's parameters.

\subsection{Fractional Kalman filtering}
\label{ssec:kalmanFilt}

The fractional-order Kalman filtering aims at estimation of the latent states at each $k$th time step with the available data. Using standard notations in the Kalman filtering (for the linear systems), we define the following estimates
\begin{equation}
\begin{split}
\hat{z}_{k} &= \mathbb{E}[z_{k}\vert x_{1},x_{2},\hdots,x_{k+1},u_{1},\hdots,u_{k}], \text{and}\\
\tilde{z}_{k} &= \mathbb{E}[z_{k}\vert x_{1},x_{2},\hdots,x_{k},u_{1},\hdots,u_{k-1}].
\end{split}
\label{eqn:KalmanStates}
\end{equation}

\begin{figure}
	\centering
	\begin{tikzpicture}
	\node[anchor=north west,inner sep=0] at (0,0) {\includegraphics*[viewport=0 0 270 130, width = 3in, height = 1.6in]{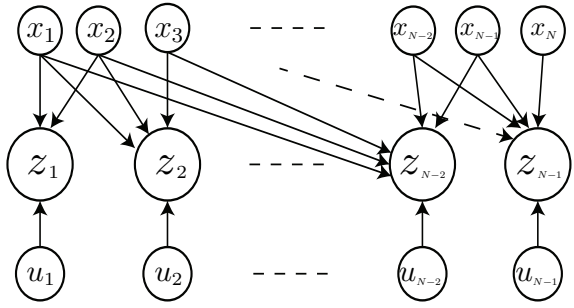}};
	\node[anchor=north west] at (0.15,-0.25) {\large{$x_{1}$}};
	\node[anchor=north west] at (0.95,-0.25) {\large{$x_{2}$}};
	\node[anchor=north west] at (1.85,-0.25) {\large{$x_{3}$}};
	\node[anchor=north west] at (5.02,-0.25) {$x_{\scaleto{N-2}{3pt}}$};
	\node[anchor=north west] at (5.91,-0.25) {$x_{\scaleto{N-1}{3pt}}$};
	\node[anchor=north west] at (6.85,-0.25) {$x_{\scaleto{N}{3pt}}$};
	
	\node[anchor=north west] at (0.15,-2) {${\scaleto{z}{7pt}}_{1}$};
	\node[anchor=north west] at (1.85,-2) {${\scaleto{z}{7pt}}_{2}$};
	\node[anchor=north west] at (5.1,-2) {{${\scaleto{z}{7pt}}_{\scaleto{N-2}{3pt}}$}};
	\node[anchor=north west] at (6.65,-2) {{${\scaleto{z}{7pt}}_{\scaleto{N-1}{3pt}}$}};
	
	\node[anchor=north west] at (0.15,-3.5) {${\scaleto{u}{5pt}}_{1}$};
	\node[anchor=north west] at (1.85,-3.5) {${\scaleto{u}{5pt}}_{2}$};
	\node[anchor=north west] at (5.1,-3.5) {${\scaleto{u}{5pt}}_{\scaleto{N-2}{3pt}}$};
	\node[anchor=north west] at (6.65,-3.5) {${\scaleto{u}{5pt}}_{\scaleto{N-1}{3pt}}$};
	\end{tikzpicture}
	\caption{Bayesian network assumption for fractional Kalman filtering.}
	\label{fig:bayesNet}
\end{figure}

\begin{figure*}
	\setcounter{equation}{3}
	\begin{eqnarray}
	\begin{bmatrix}
	A_{11}^{(t+1)\,T}\vphantom{	\sum\limits_{k=1}^{N}x_{k-1}x_{k-1}^{T}}\\
	A_{12}^{(t+1)\,T}\vphantom{\sum\limits_{k=1}^{N}(\hat{P}_{k-1}^{(t-1)} + \hat{z}^{(t-1)}\hat{z}^{(t-1)\,T})}\\
	B_{1}^{(t+1)\,T}\vphantom{\sum\limits_{k=1}^{N}u_{k-1}u_{k-1}^{T\,(t-1)}}
	\end{bmatrix} &=& 
	\begin{bmatrix}
	\sum\limits_{k=1}^{N}x_{k-1}x_{k-1}^{T} & \sum\limits_{k=1}^{N}x_{k-1}\hat{z}_{k-1}^{(t)\,T} & \sum\limits_{k=1}^{N}x_{k-1}u_{k-1}^{(t)\,T}\\
	\sum\limits_{k=1}^{N}\hat{z}_{k-1}^{(t)}x_{k-1}^{T} & \sum\limits_{k=1}^{N}(\hat{P}_{k-1}^{(t)} + \hat{z}_{k-1}^{(t)}\hat{z}_{k-1}^{(t)\,T}) & \sum\limits_{k=1}^{N}\hat{z}_{k-1}^{(t)}u_{k-1}^{T} \\
	\sum\limits_{k=1}^{N}u_{k-1}^{(t)}x_{k-1}^{T} & \sum\limits_{k=1}^{N}u_{k-1}^{(t)}\hat{z}_{k-1}^{(t)\,T} & \sum\limits_{k=1}^{N}u_{k-1}^{(t)}u_{k-1}^{(t)\,T}\\
	\end{bmatrix}^{-1}
	\begin{bmatrix}
	\sum\limits_{k=1}^{N}x_{k-1}\mathring{x}_{k}^{T}\vphantom{	\sum\limits_{k=1}^{N}x_{k-1}x_{k-1}^{T}}\\
	\sum\limits_{k=1}\hat{z}_{k-1}^{(t)}\mathring{x}_{k}^{T}\vphantom{\sum\limits_{k=1}^{N}(\hat{P}_{k-1}^{(t)} + \hat{z}^{(t-1)\,T}\hat{z}^{(t-1)\,T})}\\
	\sum\limits_{k=1}^{N}u_{k-1}^{(t)}\mathring{x}_{k}^{T}\vphantom{\sum\limits_{k=1}^{N}u_{k-1}u_{k-1}^{T\,(t)}}
	\end{bmatrix}
	\label{eqn:matUpdate1}
	\\
	\begin{bmatrix}
	A_{21}^{(t+1)\,T}\vphantom{	\sum\limits_{k=1}^{N}x_{k-1}x_{k-1}^{T}}\\
	A_{22}^{(t+1)\,T}\vphantom{\sum\limits_{k=1}^{N}(\hat{P}_{k-1}^{(t-1)} + \hat{z}^{(t-1)}\hat{z}^{(t-1)\,T})}\\
	B_{2}^{(t+1)\,T}\vphantom{\sum\limits_{k=1}^{N}u_{k-1}u_{k-1}^{T\,(t-1)}}
	\end{bmatrix}
	&=&
	\begin{bmatrix}
	\sum\limits_{k=1}^{N-1}x_{k-1}x_{k-1}^{T} & \sum\limits_{k=1}^{N-1}x_{k-1}\hat{z}_{k-1}^{(t)\,T} & \sum\limits_{k=1}^{N-1}x_{k-1}u_{k-1}^{(t)\,T}\\
	\sum\limits_{k=1}^{N-1}\hat{z}_{k-1}^{(t)}x_{k-1}^{T} & \sum\limits_{k=1}^{N-1}(\hat{P}_{k-1}^{(t)} + \hat{z}_{k-1}^{(t)}\hat{z}_{k-1}^{(t)\,T}) & \sum\limits_{k=1}^{N-1}\hat{z}_{k-1}^{(t)}u_{k-1}^{(t)\,T} \\
	\sum\limits_{k=1}^{N-1}u_{k-1}^{(t)}x_{k-1}^{T} & \sum\limits_{k=1}^{N-1}u_{k-1}^{(t)}\hat{z}_{k-1}^{T\,(t)} & \sum\limits_{k=1}^{N-1}u_{k-1}^{(t)}u_{k-1}^{T\,(t)}\\
	\end{bmatrix}^{-1}
	\begin{bmatrix}
	\sum\limits_{k=1}^{N-1}x_{k-1}\mathring{z}_{k}^{(t)\,T}\vphantom{	\sum\limits_{k=1}^{N}x_{k-1}x_{k-1}^{T}}\\
	\sum\limits_{k=1}^{N-1}(\hat{P}_{k-1}^{(t)}\Psi_{1}^{2\,T}+\hat{z}_{k-1}^{(t)}\mathring{z}_{k}^{(t)\,T})\vphantom{\sum\limits_{k=1}^{N}(\hat{P}_{k-1}^{(t-1)} + \hat{z}^{(t-1)}\hat{z}^{(t-1)\,T})}\\
	\sum\limits_{k=1}^{N-1}u_{k-1}^{(t)}\mathring{z}_{k}^{(t)\,T}\vphantom{\sum\limits_{k=1}^{N}u_{k-1}u_{k-1}^{T\,(t-1)}}
	\end{bmatrix}
	\label{eqn:matUpdate2}
	\end{eqnarray}
\end{figure*}

In contrast to the Kalman filtering for classical linear system, we have the $\hat{z}_{k}$'s conditioned on the observed data from $x_{1}$ till $x_{k+1}$. The reasoning behind this can be quickly seen in the definition of system model in equation (\ref{eqn:sysModel}). In the classical linear system, the observations and latent activities are indexed at the same time. While in the considered system model, with the observations being $x_{k}$ and latent nodes being $z_{k}$, we can witness that the equation (\ref{eqn:sysModel}) relate the latent node activity $z_{k}$ with observations till $x_{k+1}$. The Kalman filtering solutions can be mathematically intractable due to the complexities introduced by long-range dependence of the fractional operator $\Delta^{\alpha}$. We will resort to the Bayesian network assumption (as depicted in Figure\,\ref{fig:bayesNet}) for the latent state estimates which is described in the following lemma.
\begin{lemma}
	The Fractional-order Kalman filtering solution for the system described in (\ref{eqn:sysModel}) with the Bayesian Network assumption of Figure\,\ref{fig:bayesNet} is written as
	\begin{eqnarray*}
	\hat{z}_{k} &=& \tilde{z}_{k} + K_{k}(y_{k} - A_{12}^{T}\tilde{z}_{k}),\\
	\hat{P}_{k} &=& (A_{12}^{T}\Sigma_{1}^{-1}A_{12} + \tilde{P}_{k}^{-1})^{-1},\\
	K_{k} &=& \tilde{P}_{k}A_{12}^{T}(\Sigma_{1}+A_{12}\tilde{P}_{k}A_{12}^{T})^{-1},\\
	y_{k} &=& x_{k+1} + \sum\limits_{j=0}^{k+1}\Psi_{j}^{1}x_{k+1-j} - A_{11}x_{k} - B_{1}u_{k},\\
	\tilde{z}_{k} &=& A_{22}\hat{z}_{k-1} + A_{21}x_{k-1} + B_{2}u_{k-1} - \sum\limits_{j=0}^{k}\Psi_{j}^{2}\hat{z}_{k-j},\\
	\tilde{P}_{k} &=& (A_{22}-\Psi_{1}^{2})\hat{P}_{k-1}(A_{22}-\Psi_{1}^{2})^{T} \nonumber\\ 
	&&{+}\: \sum\limits_{j=2}^{k}\Psi_{j}^{2}\hat{P}_{k-j}\Psi_{j}^{2\,T} + \Sigma_{2},
	\end{eqnarray*}
\label{lemm:fracKalman}
\vspace*{-15pt}
\end{lemma}
\noindent where the conditional covariances are defined as $\hat{P}_{k} = \mathbb{E}[(z_{k}-\hat{z}_{k})(z_{k}-\hat{z}_{k})^{T}\vert x_{1},\hdots,x_{k+1},u_{1},\hdots,u_{k}]$ and $\tilde{P}_{k} = \mathbb{E}[(z_{k}-\tilde{z}_{k})(z_{k}-\hat{z}_{k})^{T}\vert x_{1},\hdots,x_{k},u_{1},\hdots,u_{k-1}]$.

Next, we present an algorithm to determine the system's parameters attaining maximum likelihood estimation.

\subsection{Maximum Likelihood Estimation}
\label{ssec:mle}

The MLE estimate of the system's parameters has to be performed in the presence of latent variables. We propose to use an Expectation-Maximization (EM) like algorithm. The conditional distributions of the latent fractional nodes considered are as described in the Section\,\ref{ssec:kalmanFilt}. Moreover, we have unknown unknowns in our system $u_{k}$, and this work in contrast to \cite{gauravACC2018} will make an estimation of $u_{k}$ as well as the input matrices $B_{i}$. We will define the EM algorithm to make an estimate of the latent fractional nodes activities and unknown unknowns jointly.

The EM update of the system's parameters at each iteration is performed via the following result.

\begin{theorem}
An update of the system parameters used in (\ref{eqn:sysModel}) with given $\{x_{k}\}_{1}^{N}, \{u_{k}^{(t)}\}_{1}^{N-1}$ and the initial conditions $x_{0},z_{0},u_{0},\hat{P}_{0}$, at each iteration index $t$ is (\ref{eqn:matUpdate1}), (\ref{eqn:matUpdate2}) and

\setcounter{equation}{5}
\begin{align}
\Sigma_{1}^{(t+1)} &= \frac{1}{N}\sum\limits_{k=1}^{N}\left[(\mathring{x}_{k} - A_{11}^{(t+1)}x_{k-1} - A_{12}^{(t+1)}\hat{z}_{k-1}^{(t)} \right.\nonumber\\
&\qquad{-} \left. B_{1}^{(t+1)}u_{k-1}^{(t)})\mathring{x}_{k}^{T}\vphantom{A_{12}^{(t+1)}x_{k-1}}\right],
\label{eqn:SigUpdate1}\\
\Sigma_{2}^{(t+1)}&= \frac{1}{N-1}\sum\limits_{k=1}^{N-1}\left[\hat{P}_{k}^{(t)}+\sum\limits_{j=1}^{k}\Psi_{j}^{2}\hat{P}_{k-j}^{(t)}\Psi_{j}^{2\,T} + \mathring{z}_{k}^{(t)}\mathring{z}_{k}^{(t)\,T}\right.\nonumber\\
&\left.\qquad - A_{21}^{(t+1)}x_{k-1}\mathring{z}_{k}^{(t)\,T} - A_{22}^{(t+1)}(\hat{P}_{k-1}^{(t)}\Psi_{1}^{2\,T}\right.\nonumber\\
&\left.\qquad + \hat{z}_{k-1}^{(t)}\mathring{z}_{k}^{(t)\,T}) - B_{2}^{(t+1)}u_{k-1}^{(t)}\mathring{z}_{k}^{(t)\,T}\vphantom{\sum\limits_{k=1}^{N}}\right],
\label{eqn:SigUpdate2}
\end{align}
\setcounter{equation}{7}
\label{thm:EM}
\end{theorem}
\noindent where $\mathring{x}_{k} = \sum\limits_{j=0}^{k}\Psi_{j}^{1}x_{k-j}$ and $\mathring{z}_{k}^{(t)} = \sum\limits_{j=0}^{k}\Psi_{j}^{1}\hat{z}_{k-j}^{(t)}$. 

%Now, using these results, we can write the iterative algorithm for system's parameter estimation as described in Algorithm\,\ref{alg:EM_alg}.

Now, using these results, we can write the iterative algorithm for system's parameter estimation. Intuitively, the algorithm starts with an initial guess of the system parameters, and then at each step, obtain the latent node activities using the approach described in Section\,\ref{ssec:kalmanFilt}. Further, upon estimating the incurred error through unknown unknowns, an update of the system's parameters is obtained and the process is repeated until convergence. The mentioned steps are formally detailed as Algorithm\,\ref{alg:EM_alg}.
%
%----------------------------------------------------------------------------
%--------------------------ALG-I : EM ALGORITHM------------------------------
%----------------------------------------------------------------------------
%

\begin{algorithm}
	{\small
		\SetKw{KwInitialize}{Initialize}
		\SetArgSty{normal}
		\KwIn{ $x[k], k \in [1,N]$ and $\alpha^{o},\alpha^{l}$}
		
		\KwOut{$\Theta = \{A_{11},A_{12},A_{21},A_{22},B_{1},B_{2},\Sigma_{1},\Sigma_{2}\}$, and $\{\hat{z}_{k}\}_{1}^{N-1}, \{u_{k}\}_{1}^{N-1}$}
		
		\KwInitialize{$x_{0},z_{0},u_{0},\hat{P}_{0}$}. For $t=0$, set $\Theta^{(0)}$ and $\{u_{k}^{(0)}\}_{1}^{N-1}$

		\Repeat{until converge}{

			\textbf{`E-step'}
			
			(i) For $k \in [1,N-1]$ obtain $\hat{z}_{k}^{(t+1)}$ from Lemma\,\ref{lemm:fracKalman} using $\{u_{k}^{(t)}\}_{1}^{N-1}$ and $\Theta^{(t-1)}$\;
			
			(ii) For $k \in [1,N-1]$ obtain $u_{k}^{(t+1)}$ as
			\begin{equation*}
			u_{k}^{(t+1)} = \text{arg}\min\limits_{u}v_{1}\Sigma_{1}^{(t)\,-1}v_{1}^{T} + v_{2}\Sigma_{2}^{(t)\,-1}v_{2}^{T} + \lambda\vert\vert u\vert\vert_{1},
			\end{equation*}
			\noindent where,
			\begin{eqnarray*}
			v_{1} &\leftarrow& \mathring{x}_{k} - A_{11}^{(t)}x_{k-1}-A_{12}\hat{z}_{k-1}^{(t+1)}-B_{1}^{(t)}u,\\
			v_{2} &\leftarrow& \mathring{z}_{k}^{(t+1)} - A_{21}^{(t)}x_{k-1} - A_{22}^{(t)}\hat{z}_{k-1}^{(t+1)} - B_{2}^{(t)}u;
			\end{eqnarray*}
			 \\
			(ii) \textbf{`M-step'}: \\ obtain $\Theta^{(t+1)}$ from Theorem\,\ref{thm:EM}\;
			
			$l \leftarrow l + 1$\;
		}
	}
	\caption{EM algorithm}
	\label{alg:EM_alg}
\end{algorithm}
%--------------------------------------------------------------------------- 
%
The proof of correctness of the Algorithm\,\ref{alg:EM_alg} follows from the Theorem\,\ref{thm:EM}, and further details of formulation can be found in the Appendix. In the next section, we discuss some applications of the proposed algorithm on variety of data.

%
%----------------------------------------------------------------------------
%-----------------------------SEC-V : EXPERIMENTS ---------------------------
%----------------------------------------------------------------------------
%
\section{Experiments}
\label{sec:experi}

\begin{figure}
	\centering
	\scalebox{0.75}{
		\begin{tikzpicture}[scale = 1.4]
		\node[anchor=north west,inner sep=0] at (0,0) {\includegraphics*[viewport=0 0 520 500, width = 3.35in, height = 3in]{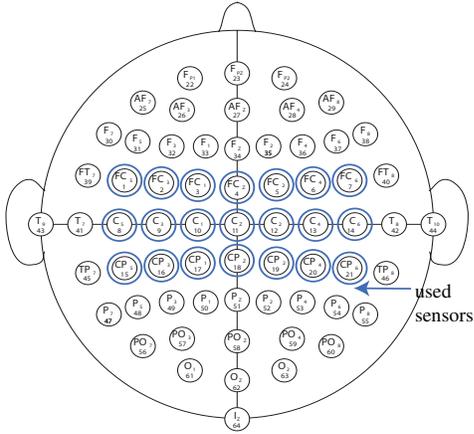}};
		\node[anchor=north west] at (5.1,-3.5) {\parbox{2cm}{used \\ sensors}};
		\end{tikzpicture}
	}
	\caption{EEG Sensor montage for the collection of the measured data (i.e., the channels). The channel labels are shown with their corresponding number.}
	\label{fig:usedSensors}
\end{figure}

We evaluate the performance of the proposed approach on a variety of datasets. First, we demonstrate on an artificially generated fractional-order time-series (see Section\,\ref{ssec:Experi_simul}). Next, we use real-world physiological signals available in the form of $64$-channel EEG time-series (see Section\,\ref{ssec:Experi_real}). The $64$-channel electrode distribution is shown in Figure\,\ref{fig:usedSensors}. The subjects were asked to perform various motor (actual and imagery) tasks, for example, left/right hand or feet, both hand or feet. The data was collected by BCI$2000$ system with sampling rate of $160$Hz \cite{schalk,goldberger}.

\subsection{Simulated Data}
\label{ssec:Experi_simul}
We consider a pedagogical fractional-order system with three nodes, and without unknown inputs with the following parameters:
\begin{eqnarray*}
A = \begin{bmatrix}
	0 & 0.1 & 0.2 \\
	-0.01 & -0.02 & 0.3\\
	0.01 & -0.03 & -0.05
\end{bmatrix}, \alpha = \{0.7, 1.1, 0.8\}.
\end{eqnarray*}

For studying the latent node behavior, we remove one node and observe the rest, $n=2$ and $m=1$, from which we use to recover the time-series generated by the accessible nodes (i.e., the nodes that are not latent) using our proposed method (i.e., with latent variables) versus those previously used in the literature \cite{gauravACC2018} (i.e., without latent variables). The five-step prediction error results are summarized in Table\,\ref{tab:Simresults}, where the relative error values are computed as 
\begin{eqnarray}
e_{i} = \sqrt{\sum\limits_{k=1}^{N}(x_{i}[k]-\hat{x}_{i}[k])^{2}\,/ \sum\limits_{k=1}^{N}x_{i}^{2}[k]}\, ,
\label{eqn:errDef}
\end{eqnarray}

%\newcolumntype{Y}{>{\centering\arraybackslash}X}
\newcolumntype{P}[1]{>{\centering\arraybackslash}p{#1}}
\begin{table}[ht]
	\centering
	\begin{tabular}{|P{0.7cm}|P{0.7cm}|c|c|c|c|}
%	\begin{tabular}{*{6}|c|}
		\hline
		\multicolumn{2}{|c}{Observed} & \multicolumn{2}{|c|}{without latent} & \multicolumn{2}{c|}{with latent} \\
		\hline
		\hline
		2 & $\color{blue}{3}$ & 13.40\% & $\color{blue}{71.14\%}$ & 7.94\% & $\color{blue}{68.22\%}$ \\
		\hline
		1 & $\color{blue}{3}$ & 22.12\% & $\color{blue}{69.26\%}$ & 21.12\% & $\color{blue}{65.31\%}$ \\
		\hline
		1 & $\color{blue}{2}$ & 11.44 \% & $\color{blue}{28.92\%}$ & 8.37\% & $\color{blue}{21.03\%}$ \\
		\hline
	\end{tabular}
	\caption{A comparison of mean squared prediction error with and without using latent node model for various possibilities of observed nodes. The $i$th row corresponds to making node\,$i$ as latent.}
	\label{tab:Simresults}
\end{table}

\begin{figure}
	\centering
	\includegraphics*[viewport=40 0 675 340, width = 3.35in, height = 1.7in]{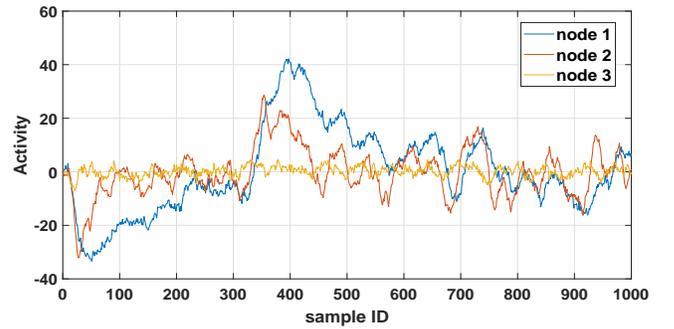}
\caption{Simulated activities for all three nodes generated by selected model parameters.}
\label{fig:simData}
\end{figure}
\noindent where $\hat{x}_{i}[k]$ is the predicted value of the $i$th node at time $k$. The error percentage is consistently high for node\,$3$, and the reason for this lies in the actual behavior of the node activity as seen in Figure\,\ref{fig:simData}. Specifically, node\,$3$ activity --unlike other two nodes-- stays very close to zero and vary frequently, which makes it difficult to use for accurate predictions using the proposed model. Next, we see the application on real-world EEG dataset.
\newcolumntype{Y}{>{\centering\arraybackslash}X}
\begin{table*}[t]
	\centering
	\begin{subtable}{\textwidth}
		\begin{tabularx}{\textwidth}{|c|c|Y|Y|Y|Y|Y|Y|Y|Y|}
		\hline
		Observed (1$\rightarrow$12) + & $\phi$ & 13 &  13$\rightarrow$14& 13$\rightarrow$15& 13$\rightarrow$16& 13$\rightarrow$17& 13$\rightarrow$18& 13$\rightarrow$19& 13$\rightarrow$20 \\
		\hline
		Hidden & 13$\rightarrow$21 & 14$\rightarrow$21& 15$\rightarrow$21&  16$\rightarrow$21&  17$\rightarrow$21&  18$\rightarrow$21&  19$\rightarrow$21& 20$\rightarrow$21& 21 \\
		\hline\hline
		Without latent (in \%) & 10.51 &	10.44  & 10.97  & 11.71  & 13.51  & 13.77  & 15.29  & 16.03  & 14.56 \\
		\hline
		With latent (in \%) & 6.07  & 7.20  & 6.53  & 8.49  & 11.36  & 10.50 & 13.18  & 13.18  & 13.16  \\
		\hline
	\end{tabularx}
	\vspace*{-4pt}
	\caption{}
	\label{tab:EEG1}
	\end{subtable}
	\begin{subtable}{\textwidth}
		\begin{tabularx}{\textwidth}{|c|Y|Y|Y|Y|Y|Y|Y|Y|Y|}
		\hline
		Observed (1$\rightarrow$12) + & $\phi$ & 56 &  56$\rightarrow$57& 56$\rightarrow$58& 56$\rightarrow$59& 56$\rightarrow$60& 56$\rightarrow$61& 56$\rightarrow$62& 56$\rightarrow$63 \\
		\hline
		Hidden & 56$\rightarrow$64 & 57$\rightarrow$64& 58$\rightarrow$64&  59$\rightarrow$64&  60$\rightarrow$64&  61$\rightarrow$64&  62$\rightarrow$64& 63$\rightarrow$64& 64 \\
		\hline\hline
		Without latent (in \%) & 10.51 & 12.09 & 15.99 & 13.45 & 15.29 &  16.59 & 20.40 & 20.89 & 21.52 \\
		\hline
		With latent (in \%) & 6.31 & 7.92 & 11.27 & 9.25 & 9.90 &  9.69 & 13.28 & 19.87 & 20.07  \\
		\hline
	\end{tabularx}
	\caption{}
	\label{tab:EEG2}
	\end{subtable}
\vspace*{-7pt}
\caption{Average prediction error with and without using latent model for two different set of sensors in (a) and (b). Each column has labeled hidden nodes, and observed nodes are union of $(1\rightarrow 12)$ and the corresponding column entry. The total number of observed and latent nodes change by $n+1$ and $m-1$ from left to right.
}
\label{tab:EEGBase}
\end{table*}

\subsection{Real-world data}
\label{ssec:Experi_real}

%The physiological signals like EEG datasets are well modeled with high prediction accuracies using existing time-varying complex networks with fractional dynamics \cite{xue1, gauravACC2018, gauravICCPS2018}.

\begin{figure}
	\centering
	\begin{subfigure}{\linewidth}
	\includegraphics*[viewport=0 0 565 350, width = 3.0in, height = 1.7in]{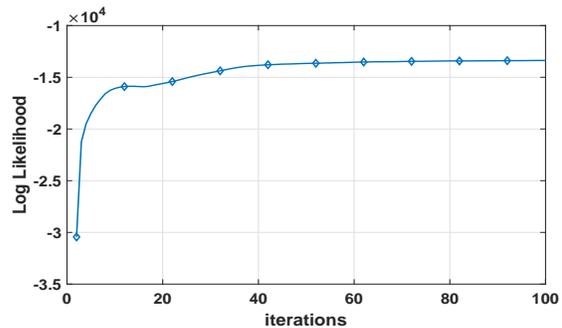}
	\caption{}
	\end{subfigure}
\\
\begin{subfigure}{\linewidth}
	\includegraphics*[viewport=0 0 565 350, width = 3.0in, height = 1.7in]{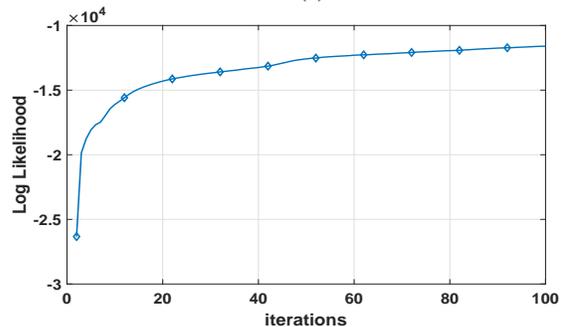}
	\caption{}
\end{subfigure}
	\caption{Log likelihood vs number of iterations for observed indices $(1\rightarrow 12)$, and hidden indices $(13\rightarrow 21)$ in (a) and  $(56\rightarrow 64)$ in (b), using Algorithm\,\ref{alg:EM_alg}.}
	\label{fig:ll_1}
\end{figure}

The estimation of model's parameters, such as the coupling matrices $A_{ij}$, the input matrices $B_{i}$, and the latent states, is performed for the EEG data using Algorithm\,\ref{alg:EM_alg}. The log likelihood converges with iterations as we observe in Figure\,\ref{fig:ll_1}. We notice that the choice of initial conditions can play a great role in fast convergence, as well as the accuracy of the results. If some previous knowledge is available (for example, through experiments) about the coupling matrices, then it can be used to achieve better results. In this work, the matrices are initialized to entries selected uniformly at random between $-1$ and $1$. In the current experiment, we have used the data when the subject has executed `both feet movement'. While performing model estimation, the use of relevant EEG sensors are required for having accurate predictions. Therefore, we used neuro-physiological evidence-based sensors that capture the behavior associated with the peripheral nervous system (i.e., the motor cortex) and labeled from $1$ to $21$, --see Figure\,\ref{fig:usedSensors}. Specifically, different subregions in the motor region are activated when the feet move, so we have used this information to carefully reduce the number of sensors/nodes in our study.

The proposed latent model is tested in a comprehensive manner by performing the following steps: \emph{(i)} first fixing the nodes to make prediction from sensor IDs $1$ to $12$ (denoted by $1\rightarrow 12$); \emph{(ii)} second, consecutively reveal new nodes to increase the total observed nodes dimension $n$ (originally from $12$) by one and decrease total latent node dimension $m$ (from $9$) by one, in each step. The reported error values are computed from equation (\ref{eqn:errDef}), and are averaged across the fixed twelve observed nodes. In Table\,\ref{tab:EEG1} we provide some evidence that the latent model with minimal information concerning fractional orders of the latent nodes perform better than without using any model on the unobserved nodes. We also observed the necessity of the relevant latent nodes, by considering the set of sensors which are placed on the region of brain least related to the undertaken situation of `both feet movement'. The same experiment is repeated to predict the activities of fixed nodes in consideration $1\rightarrow 12$ and varying the total observed/latent nodes, but this time from a set of sensor IDs $\{56,\hdots,64\}$. The prediction error values are reported in Table\,\ref{tab:EEG2}. We notice that the error values are higher upon revealing nodes from the set $\{56,\hdots,64\}$. This raises an important and intuitive point that, revealing/hiding \mbox{time-series} that have less relation to the data under consideration are very likely to increase inaccuracies in the model.

The experiments, both simulated and with real EEG data, provided evidence that the inclusion of latent model is helpful in the context of the accuracy of the retrieved model and prediction accuracies.

%The experiments, both simulated and with real data, demonstrated that proposed latent model can help in better understanding the data in hand, by accepting the fact that there exist nodes in the network which are not probed. The relevance of single parameter, fractional-order, is shown by using it as available information to predict the activities of latent nodes, and improving model of the observed nodes.

%
%----------------------------------------------------------------------------
%-----------------------------SEC-VI : CONCLUSION----------------------------
%----------------------------------------------------------------------------
%

\section{Conclusion}
\label{sec:concl}

We have introduced and studied the framework of latent nodes in the TVCN with fractional dynamical model and additional unknown drivers. An iterative solution is proposed which jointly estimate the system's parameters, the latent node states as they evolve over time, and the unknown unknowns (i.e., the unknown input matrices and inputs). We have shown that the minimal assumption of knowledge of fractional coefficients of the latent node allows us to better explain the observed data. We have applied the proposed concepts on both simulated and experimental data to see the gains in prediction accuracy of the observed data.

Future work will focus on the intriguing role of fractional-order coefficients of the latent nodes. Instead of the entire latent node data, it is observed that the associated fractional-order coefficients are enough for removing the effects of hidden nodes. Moreover, the fractional coefficients can be used to decide the required dimensions, or degrees-of-freedom, of the hidden part to make observed and hidden a complete system.

\section{Acknowledgment}
The authors are thankful to the reviewers for their valuable feedback. G.G. and P.B. gratefully acknowledge the support by the U.S. Army Defense Advanced Research Projects Agency (DARPA) under grant no. W911NF-17-1-0076, the DARPA Young Faculty Award under grant no. N66001-17-1-4044 support, and the National Science Foundation Career award under grant no. CPS/CNS-1453860. The views, opinions, and/or findings contained in this article are those of the authors and should not be interpreted as representing the official views or policies, either expressed or implied by the Defense Advanced Research Projects Agency, the Department of Defense or the National Science Foundation.

%
%----------------------------------------------------------------------------
%--------------------------------APPENDIX------------------------------------
%----------------------------------------------------------------------------
%
{
\appendices
\section{Proof of Lemma\,\ref{lemm:fracKalman}}

\begin{proof}
With the assumption of initial states $x_{0}$ and $z_{0}$ being normal distributed, and $\hat{P}_{0}$ known, the Kalman filtering is application of the Bayes' formula.
\begin{eqnarray}
\text{log}~P(z_{k}\vert x_{0},x_{1},\hdots,x_{k+1},u_{1},\hdots,u_{k}) = \nonumber\\
 \quad\text{log}~P(x_{k+1}\vert x_{0},x_{1},\hdots,z_{k}) \nonumber\\
{+}\,\text{log}~P(z_{k}\vert x_{0},x_{1},\hdots,x_{k}) + \hdots
\label{eqn:bayesProb}
\end{eqnarray}
Let $\mathcal{H}_{k} = \{x_{0},x_{1},\hdots,x_{k+1},u_{1},\hdots,u_{k}\}$, $z_{k}\vert \mathcal{H}_{k}\sim\mathcal{N}(\hat{z},\hat{P}_{k})$, and $z_{k}\vert \mathcal{H}_{k-1}\sim\mathcal{N}(\tilde{z},\tilde{P}_{k})$. Upon comparing the terms within equation (\ref{eqn:bayesProb}) and using (\ref{eqn:sysModel}), we can write
\begin{eqnarray}
\hat{P}_{k}^{-1} &=& A_{12}^{T}\Sigma_{1}^{-1}A_{12} + \tilde{P}_{k}^{-1},\nonumber\\
\hat{z}_{k} &=& \tilde{z}_{k} + K_{k}(x_{k+1} + \sum\limits_{j=0}^{k+1}\Psi_{j}^{1}x_{k+1-j} \nonumber\\
&&{-}\: A_{11}x_{k} - B_{1}u_{k} - A_{12}^{T}\tilde{z}_{k}),
\end{eqnarray}
\noindent where $K_{k} = \tilde{P}_{k}A_{12}^{T}(\Sigma_{1}+A_{12}\tilde{P}_{k}A_{12}^{T})^{-1}$ using matrix inversion lemma \cite{Golub1996}. For $\tilde{z}_{k}$, we can write
\begin{align}
\tilde{z}_{k} &= \mathbb{E}[z_{k}\vert \mathcal{H}_{k-1}]\nonumber\\
&= \mathbb{E}[\mathbb{E}[z_{k}\vert z_{0},z_{1},\hdots,z_{k-1},x_{k-1}]\vert \mathcal{H}_{k-1}]\nonumber\\
&\stackrel{(a)}{=}\mathbb{E}[A_{22}z_{k-1} + A_{21}x_{k-1} + B_{2}u_{k-1} \nonumber\\
&\qquad{-}\:\sum\limits_{j=1}^{k}\Psi_{j}^{2}z_{k-j}\vert \mathcal{H}_{k-1}]\nonumber\\
&\stackrel{(b)}{=} A_{22}\hat{z}_{k-1} + A_{21}x_{k-1} + B_{2}u_{k-1} - \sum\limits_{j=0}^{k}\Psi_{j}^{2}\hat{z}_{k-j},
\label{eqn:zHatEqn}
\end{align}
\noindent where $(a)$ follows from equation (\ref{eqn:sysModel}), and $(b)$ is obtained using Bayesian network assumption of Figure\,\ref{fig:bayesNet}. To obtain the recursion for covariance matrix $\tilde{P}_{k}$, we begin by expressing $z_{k}-\tilde{z}_{k}$ using (\ref{eqn:zHatEqn}) and (\ref{eqn:sysModel}) as
\begin{eqnarray}
z_{k} - \hat{z}_{k} &=& (A_{22}-\Psi_{1}^{2})(z_{k-1}-\hat{z}_{k-1}) \nonumber\\
&&{-}\: \sum\limits_{j=2}^{k}\Psi_{j}^{2}(z_{k-j}-\hat{z}_{k-j}) + e_{2,k}.
\label{eqn:zHatDiff}
\end{eqnarray}

Using (\ref{eqn:zHatDiff}), the covariance matrix $\tilde{P}_{k}$ is written as
\begin{align}
\tilde{P}_{k} &= (A_{22}-\Psi_{1}^{2})\hat{P}_{k-1}(A_{22}-\Psi_{1}^{2})^{T} + \Sigma_{2}\nonumber\\
&\qquad{+} \sum\limits_{j=2}^{k}\Psi_{j}^{2}\hat{P}_{k-j}\Psi_{j}^{2\,T}- \sum\limits_{j=2}^{k}(A_{22}-\Psi_{1}^{2}) \nonumber\\
&\quad\mathbb{E}[(z_{k-1}-\hat{z}_{k-1})(z_{k-j}-\hat{z}_{k-j})^{T}\vert \mathcal{H}_{k-1}]\Psi_{j}^{2\,T}\nonumber\\
&\quad{-}\sum\limits_{j=2}^{k}\Psi_{j}^{2}\mathbb{E}[(z_{k-j}-\hat{z}_{k-j})(z_{k-1}-\hat{z}_{k-1})^{T}\vert \mathcal{H}_{k-1}]\nonumber\\
&\qquad(A_{22}-\Psi_{1}^{2})^{T} + \sum\limits_{i>j\geq 2}\Psi_{i}^{2}\mathbb{E}[(z_{k-i}-\hat{z}_{k-j})\nonumber\\
&\qquad (z_{k-i}-\hat{z}_{k-j})^{T}\vert \mathcal{H}_{k-1}]\Psi_{j}^{2\,T}.
\label{eqn:PHatFull}
\end{align}
From the Bayesian network assumption of Figure\,\ref{fig:bayesNet}, it follows that
\begin{equation}
\mathbb{E}[(z_{n}-\hat{z}_{n})(z_{m}-\hat{z}_{m})^{T}\vert \mathcal{H}_{l}] = 0,\quad \forall n\neq m \leq l.
\label{eqn:crossCovBayesian}
\end{equation}
Therefore, using (\ref{eqn:crossCovBayesian}), equation (\ref{eqn:PHatFull}) reduces to 
\begin{eqnarray}
\tilde{P}_{k} &=& (A_{22}-\Psi_{1}^{2})\hat{P}_{k-1}(A_{22}-\Psi_{1}^{2})^{T} \nonumber\\ 
&&{+}\: \sum\limits_{j=2}^{k}\Psi_{j}^{2}\hat{P}_{k-j}\Psi_{j}^{2\,T} + \Sigma_{2}.
\end{eqnarray}
\end{proof}

\section{Proof of Theorem\,\ref{thm:EM}}
\begin{proof}
Let, $\{x_{k}\}_{1}^{N}, \{u_{k}\}_{1}^{N-1}$ and the initial conditions $x_{0}, z_{0}, u_{0}, \hat{P}_{0}$ be known. Let us also denote the set of variables $\mathcal{H}_{k} = \{x_{0},x_{1},\hdots,x_{k+1},u_{0},u_{1},\hdots,u_{k}\}$. The `$Q$ function' for the EM like algorithm (described in Algorithm\,\ref{alg:EM_alg}) with latent variables being $z_{k}$ can be written as
\begin{align}
&Q(\Theta ;\Theta^{(t)}) \nonumber\\
&=\mathbb{E}_{z_{1},\hdots,z_{N-1}\vert \mathcal{H}_{N-1};\Theta^{(t)}}[\text{log}P(z_{0},z_{1},\hdots,z_{N-1},\mathcal{H}_{N-1};\Theta)]
\nonumber\\
&\stackrel{(a)}{=}\mathbb{E}_{z_{1}\vert \mathcal{H}_{1};\Theta^{(t)}}\mathbb{E}_{z_{2}\vert \mathcal{H}_{2};\Theta^{(t)}}\hdots\mathbb{E}_{z_{N-1}\vert \mathcal{H}_{N-1};\Theta^{(t)}}[\nonumber\\
&\qquad\text{log}P(z_{0},z_{1},\hdots,z_{N-1},\mathcal{H}_{N-1};\Theta)]\nonumber\\
&\stackrel{(b)}{=}\mathbb{E}_{z_{1}\vert \mathcal{H}_{1};\Theta^{(t)}}\mathbb{E}_{z_{2}\vert \mathcal{H}_{2};\Theta^{(t)}}\hdots\mathbb{E}_{z_{N-1}\vert \mathcal{H}_{N-1};\Theta^{(t)}}[\nonumber\\
&\qquad \sum\limits_{k=1}^{N-1}\text{log}P(z_{k}\vert z_{0},\hdots,z_{k-1},\mathcal{H}_{N-1};\Theta)\nonumber\\
&\qquad{+}\sum\limits_{k=1}^{N}\text{log}P(x_{k}\vert z_{0},\hdots,z_{k-1},\mathcal{H}_{N-1};\Theta)],
\end{align}
\noindent where $(a)$ is written using Bayesian assumption of Figure\,\ref{fig:bayesNet}, and $(b)$ can be written due to the uncorrelated noise assumption of $e_{1}$ and $e_{2}$ as assumed in (\ref{eqn:sysModel}). For notational convenience, we have dropped the constants at each step. The terms inside the summation can be written as 
\begin{eqnarray}
\text{log}P(x_{k}\vert z_{0},\hdots,z_{k-1},\mathcal{H}_{N-1};\Theta) = -\frac{1}{2}\text{log}\vert \Sigma_{1}\vert \nonumber\\
-\frac{1}{2}e_{1\,k}^{T}\Sigma_{1}^{-1}e_{1\,k},\\
\text{log}P(z_{k}\vert z_{0},\hdots,z_{k-1},\mathcal{H}_{N-1};\Theta) = -\frac{1}{2}\text{log}\vert \Sigma_{2}\vert \nonumber\\
-\frac{1}{2}e_{2\,k}^{T}\Sigma_{2}^{-1}e_{2\,k},
\end{eqnarray}
\noindent where we have dropped the constants for notational convenience, and $e_{1\,k}, e_{2\,k}$ are from equation (\ref{eqn:sysModel}). We now proceed to obtain the $Q$ function which is described as follows:
\begin{align}
Q(\Theta;\Theta^{(t)}) &= -\frac{N}{2}\text{log}\vert\Sigma_{1}\vert-\frac{N-1}{2}\text{log}\vert\Sigma_{1}\vert\nonumber\\
&-\mathbb{E}_{z_{1}\vert \mathcal{H}_{1};\Theta^{(t)}}\mathbb{E}_{z_{2}\vert \mathcal{H}_{2};\Theta^{(t)}}\hdots\mathbb{E}_{z_{N-1}\vert \mathcal{H}_{N-1};\Theta^{(t)}}\nonumber\\
&\qquad[\sum\limits_{k=1}^{N}e_{1\,k}^{T}\Sigma_{1}^{-1}e_{1\,k} + \sum\limits_{k=1}^{N}e_{2\,k}^{T}\Sigma_{1}^{-1}e_{2\,k}].
\end{align}
Using Lemma\,\ref{lemm:fracKalman} with $\Theta^{(t)}$, $\{x_{k}\}_{1}^{N}$, and $\{u_{k}\}_{1}^{N-1}$, we have
\begin{align}
&Q(\Theta;\Theta^{(t)}) = -\frac{N}{2}\text{log}\vert\Sigma_{1}\vert-\frac{N-1}{2}\text{log}\vert\Sigma_{1}\vert\nonumber\\
&-\frac{1}{2}\text{tr}\Big(\Sigma_{1}^{-1}\sum\limits_{k=1}^{N}(\mathring{x}_{k}\mathring{x}_{k}^{T} - 2A_{11}x_{k-1}\mathring{x}_{k}^{T} - 2A_{12}\hat{z}_{k-1}^{(t)}\mathring{x}_{k}^{T}\nonumber\\
&- 2B_{1}u_{k-1}\mathring{x}_{k}^{T}+A_{11}x_{k-1}x_{k-1}^{T}A_{11}^{T} + 2A_{12}\hat{z}_{k-1}^{(t)}x_{k-1}^{T}A_{11}^{T}\nonumber\\
&+2B_{1}u_{k-1}x_{k-1}A_{11}^{T} + A_{12}(\hat{P}_{k-1}^{(t)}+\hat{z}_{k-1}^{(t)}\hat{z}_{k-1}^{(t)\,T})A_{12}^{T}\nonumber\\
& +2B_{1}u_{k-1}\hat{z}_{k-1}^{(t)\,T}A_{12}^{T} + B_{1}u_{k-1}u_{k-1}^{T}B_{1}^{T})\Big)\nonumber\\
&-\frac{1}{2}\text{tr}\Big(\Sigma_{2}^{-1}\sum\limits_{k=1}^{N-1}(\hat{P}_{k}^{(t)}+\sum\limits_{j=1}^{k}\Psi_{j}^{2}\hat{P}_{k-j}^{(t)}\Psi_{j}^{2\,T} + \mathring{z}^{(t)}_{k}\mathring{z}_{k}^{(t)\,T}\nonumber\\
&-2A_{21}x_{k-1}\mathring{z}_{k}^{(t)\,T} - 2A_{22}(\hat{P}_{k-1}^{(t)}\Psi_{1}^{2\,T}+\hat{z}_{k-1}^{(t)}\mathring{z}_{k}^{(t)\,T})\nonumber\\
&-2B_{2}u_{k-1}\mathring{z}_{k}^{(t)\,T} + A_{21}x_{k-1}x_{k-1}^{T}A_{21}^{T} + 2A_{21}x_{k-1}\hat{z}_{k-1}^{(t)\,T}A_{22}^{T}\nonumber\\
&+2A_{21}x_{k-1}u_{k-1}^{T}B_{2}^{T}+A_{22}(\hat{P}_{k-1}^{(t)}+\hat{z}_{k-1}^{(t)}\hat{z}_{k-1}^{(t)\,T}A_{22}^{T}\nonumber\\
&+A_{22}\hat{z}_{k-1}^{(t)}u_{k-1}^{T}B_{2}^{T} + B_{2}u_{k-1}u_{k-1}^{T}B_{2}^{T})\Big).
\label{eqn:QFunFull}
\end{align}
For the `M-step', we differentiate the $Q$ function with respect to parameters set $\Theta$ and equate it to zero. For example, upon differentiating equation (\ref{eqn:QFunFull}) with $A_{11}$, we obtain
\begin{eqnarray}
\sum\limits_{k=1}^{N}(A_{11}x_{k-1}x_{k-1}^{T} + A_{12}\hat{z}_{k-1}^{(t)}x_{k-1}^{T} + B_{1}u_{k-1}x_{k-1}^{T}) \nonumber\\
= \sum\limits_{k=1}^{N}\mathring{x}_{k}x_{k-1}^{T}.
\end{eqnarray}
Similarly, differentiating equation (\ref{eqn:QFunFull}) with respect to $A_{12},B_{1},A_{21},A_{22}$ and $B_{2}$ and setting them to zero, after properly rearranging the terms, we obtain the next iteration update as shown in (\ref{eqn:matUpdate1}) and (\ref{eqn:matUpdate2}). The result of differentiating (\ref{eqn:QFunFull}) with $\Sigma_{1}$ and $\Sigma_{2}$, and setting them to zero lead us to the equations (\ref{eqn:SigUpdate1}) and (\ref{eqn:SigUpdate2}), respectively.
\end{proof}

\section{EM formulation}
The `E-step' of the EM like algorithm consists of estimating the unknown unknowns $\{u_{k}\}_{1}^{N-1}$ and the latent fractional states $\{z_{k}\}_{1}^{N-1}$. For the estimation of unknown unknowns which do not obey the fractional dynamics, let us consider the following (at $t+1$ iteration)
\begin{equation}
u_{k}^{\ast} = \arg\max\limits_{u_{k}}P(u_{k}\vert x_{1},\hdots,x_{k+1}, \hat{z}_{1}^{(t+1)},\hdots,\hat{z}_{k+1}^{(t+1)};\Theta^{(t)}).
\label{eqn:uEst1}
\end{equation}
We can select some conjugate prior for the unknown inputs, for example, Laplacian, such that $P(u_{k})\propto \text{exp}(-\lambda\vert\vert u_{k}\vert\vert_{1})$. Therefore, equation (\ref{eqn:uEst1}) can be re-written as
\begin{align*}
u_{k}^{\ast} &= \arg\max\limits_{u_{k}}\text{log}P(u_{k}\vert x_{1},\hdots,x_{k+1}, \hat{z}_{1}^{(t+1)},\hdots,\hat{z}_{k}^{(t+1)};\Theta^{(t)})\nonumber\\
u_{k}^{\ast}&= \arg\max\limits_{u_{k}}\text{log}P({x}_{k+1}\vert x_{1},\hdots,x_{k},\hat{z}_{k}^{(t+1)}\vert u_{k};\Theta^{(t)}) \nonumber\\
&\qquad{+}\: \text{log}P(\hat{z}_{k+1}^{(t+1)}\vert x_{1},\hdots,x_{k},\hat{z}_{k}^{(t+1)}\vert u_{k};\Theta^{(t)}) \nonumber\\
&\qquad{+}\: \text{log}P(u_{k})\nonumber\\
&=\arg\max\limits_{u_{k}}-\frac{1}{2}v_{1\,k}\Sigma_{1}^{-1}v_{1,\,k} -\frac{1}{2}v_{2\,k}\Sigma_{2}^{-1}v_{2,\,k} -\lambda\vert\vert u_{k}\vert\vert_{1}
\end{align*}
\noindent where,
% $v_{1\,k} = 
\begin{eqnarray*}
v_{1\,k} &=& \mathring{x}_{k} - A_{11}^{(t)}x_{k-1}-A_{12}\hat{z}_{k-1}^{(t)}-B_{1}^{(t)}u_{k},\\
v_{2\,k} &=& \mathring{z}_{k}^{(t)} - A_{21}^{(t)}x_{k-1} - A_{22}^{(t)}\hat{z}_{k-1}^{(t)} - B_{2}^{(t)}u_{k}.
\end{eqnarray*}
We will approximate the conditional distribution of unknown inputs as $P(u_{k}\vert x_{1},\hdots,x_{k+1},\hat{z}_{1}^{(t+1)},\hdots,\hat{z}_{1}^{(t+1)})^{(t+1)} \approx \mathds{1}_{u_{k}=u_{k}^{\ast}}$. This is sometimes referred as the `hard EM'. Lastly, the rest of the `E-step' readily follows from Lemma\,\ref{lemm:fracKalman}. The `M-step' follows from Theorem\,\ref{thm:EM}.

}

%
%----------------------------------------------------------------------------
%--------------------------------REFERENCES----------------------------------
%----------------------------------------------------------------------------
%

\footnotesize

\bibliographystyle{IEEEtran}
\bibliography{IEEEabrv,hidState}

\end{document}